\documentclass[letterpaper, 10 pt, conference]{ieeeconf}
\IEEEoverridecommandlockouts
\usepackage{cite}
\usepackage{amsmath,amssymb,amsfonts}
\usepackage{algorithmic}
\usepackage{graphicx}
\usepackage{textcomp}
\usepackage{xcolor}

\usepackage{balance}
\usepackage{url}
\usepackage{hyperref}
\usepackage{siunitx}
\usepackage{multicol}
\usepackage{multirow}
\usepackage{tabularx}
\usepackage{booktabs}
\usepackage[font=small]{subcaption}
\usepackage[font={small}]{caption} 

\def\BibTeX{{\rm B\kern-.05em{\sc i\kern-.025em b}\kern-.08em
    T\kern-.1667em\lower.7ex\hbox{E}\kern-.125emX}}
    
\begin{document}

\title{\LARGE \bf Towards Multi-robot Exploration: \\A Decentralized Strategy for UAV Forest Exploration}

\author{Luca Bartolomei, Lucas Teixeira and Margarita Chli \\
Vision For Robotics Lab, ETH Z{\"u}rich, Switzerland%
\thanks{This work was supported by NCCR Robotics, the Amazon Research Awards, and the HILTI group.}%
}

\maketitle

\thispagestyle{empty}
\pagestyle{empty}

\begin{abstract}
\label{ch:abstract}
Efficient exploration strategies are vital in tasks such as search-and-rescue missions and disaster surveying.
Unmanned Aerial Vehicles (UAVs) have become particularly popular in such applications, promising to cover large areas at high speeds. 
Moreover, with the increasing maturity of onboard UAV perception, research focus has been shifting toward higher-level reasoning for single- and multi-robot missions.
However, autonomous navigation and exploration of previously unknown large spaces still constitutes an open challenge, especially when the environment is cluttered and exhibits large and frequent occlusions due to high obstacle density, as is the case of forests.
Moreover, the problem of long-distance wireless communication in such scenes can become a limiting factor, especially when automating the navigation of a UAV swarm.
In this spirit, this work proposes an exploration strategy that enables UAVs, both individually and in small swarms, to quickly explore complex scenes in a decentralized fashion.
By providing the decision-making capabilities to each UAV to switch between different execution modes, the proposed strategy strikes a great balance between cautious exploration of yet completely unknown regions and more aggressive exploration of smaller areas of unknown space.
This results in full coverage of forest areas of variable density, consistently faster than the state of the art.
Demonstrating successful deployment with a single UAV as well as a swarm of up to three UAVs, this work sets out the basic principles for multi-root exploration of cluttered scenes, with up to $\mathbf{65\%}$ speed up in the single UAV case and $\mathbf{40\%}$ increase in explored area for the same mission time in multi-UAV setups.
%
%
\end{abstract}
\section{Introduction}
\label{ch:intro}
The growing interest in Unmanned Aerial Vehicles (UAVs) has led to their extensive deployment in tasks such as inspection and search-and-rescue missions.
In these applications, the capacity of the robot to quickly explore and map unknown environments autonomously is fundamental.
%
The literature on this topic is extensive, and many different approaches have been proposed throughout the years \cite{selin2019efficient, Schmid20ActivePlanning, kompis2021informed, Cieslewski2017rapid, zhou2021fuel}.
However, one of the biggest challenges in the exploration of unknown environments is the capacity to achieve a good trade-off between the competing goals of shorter exploration times of an area of interest (i.e. pushing for high-speed navigation) and safety, which requires caps on the velocity of each robot.
In fact, navigating in the vicinity of the boundaries between known and unknown space is challenging, as the robot can get stuck in dead ends, or needs to perform complex dodging maneuvers to avoid collisions.
Consequently, to maintain the safety of both the platform and its surroundings, most path planners generate conservative start-and-stop motions, not fully exploiting the capacity of a UAV to fly at high speeds.
This effect is exacerbated when the environment to explore is particularly cluttered, as is the case in forests, leading to inefficient and incomplete coverage.
By design, these methods generally drive the exploration process by biasing exploration towards large areas of unexplored space.
While this strategy could be advantageous in open and wide spaces, it can be detrimental when exploring cluttered scenes.
In fact, the main pitfall of such strategies is that, while the exploration process attempts to cover as much unknown space as possible, when this is deployed in environments with many obstacles, thinner trails of unknown space are left unexplored (e.g. due to occlusions), imposing the need for a second sweep of the environment over mostly explored areas.
\begin{figure}[t!]
  \centering
  \includegraphics[width=\linewidth]{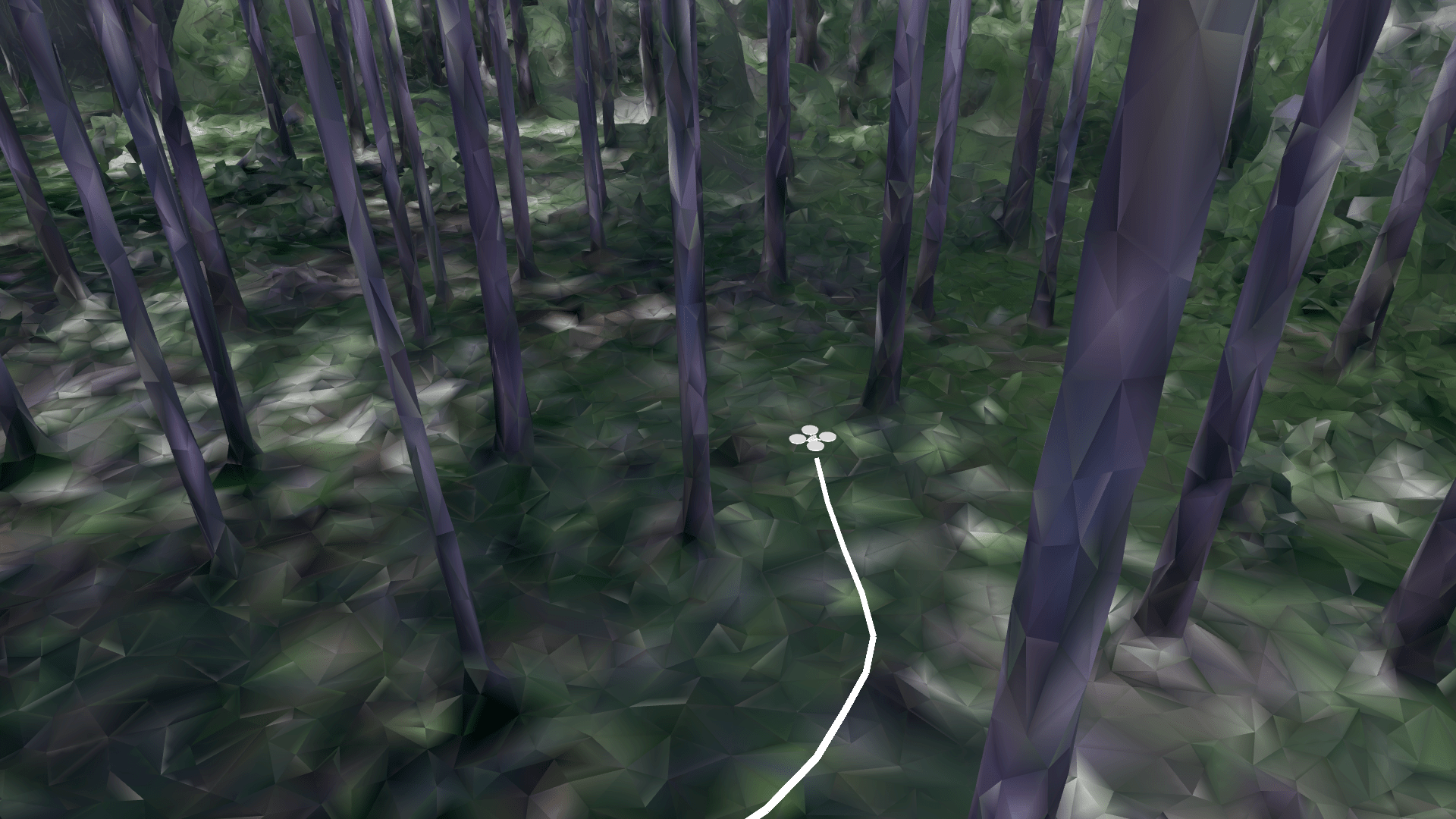}
  \caption{3D-view of the proposed system guiding safe and successful exploration of a UAV in a digital model of a real-world forest \cite{afzal2021swarm}. The planner is able to avoid collisions between the UAV and the obstacles, clearing frontiers on-the-go by balancing cautious navigation with aggressive exploitation of known, free space, in a bid to maximize the efficiency of the exploration.\vspace{-5mm}}
  \label{fig:intro:real_forest}
\end{figure}
%

%
Aiming to mitigate these issues, pushing for faster coverage of the areas of interest, multi-robot extensions for exploration have also been proposed \cite{yulun2020search, rouvcek2019darpa, bartolomei2020server, corah2019communication}.
However, these focus on the problem of coordination at the system-level and, and while they can perform better from a global planning point of view, they suffer from the same limitations as the single-UAV case in obstacle-dense environments.
Motivated by these challenges, in this work we propose an exploration strategy for autonomous UAV robots aiming to explore forests of increasing tree density, as they pose some of the most difficult challenges for exploration planning.
Our objective is to exploit the platform's dynamics to the fullest despite the high density of obstacles, in order to achieve the complete coverage of the environment efficiently.
To this end, the proposed strategy enables switching between two different behaviors for each robot; namely, cautious exploration of unknown space and more aggressive maneuvers when navigating in already explored areas to clear smaller portions of unknown space caused by occlusions.
We evaluate the proposed approach in a series of challenging experiments in simulation, both in randomly generated forests and in a 3D reconstruction of a real forest (Fig \ref{fig:intro:real_forest}). 
Benchmarking against the state of the art reveals superior efficiency for the proposed approach achieving higher overall UAV speeds and lower exploration times.
Finally, aiming to set out the scaffolding toward decentralized multi-robot exploration planning, we show how the proposed strategy can accommodate more than one UAV in the exploration mission, and we demonstrate that our method performs comparably to or better than a map-splitting centralized approach.
In summary, the contributions of this work are as follows:
\begin{itemize}
    \item the design of an exploration strategy, able to strike an effective balance between cautious exploration and aggressive exploitation of the explored map,
    \item the extension of the single-robot design to a multi-robot decentralized approach, and
    \item extensive evaluations in simulation, demonstrating better performance than the state of the art.
\end{itemize}
\section{Related Works}
\label{ch:rel_works}
Autonomous exploration of unknown environments with UAVs has been an active field of research over the past few decades.
The most popular approach to exploring an area of interest is to use frontiers, defined as the boundary between known and unknown space \cite{Yamauchi1997}.
These can be utilized to identify potentially informative spatial regions in order to drive the exploration process efficiently until no new frontiers are found and the exploration process can be considered complete.
There are different criteria used to decide which frontier to explore next, such as their proximity to the current field of view, following a greedy selection strategy, or having global planning dictate the selection \cite{dasilva2020}.
However, while frontier-based approaches have been proven to yield satisfactory performances, especially in terms of coverage \cite{Schmid20ActivePlanning, kompis2021informed}, they generally lead to inefficient motions and sub-optimal action selection.
This is mostly caused by the sensing modalities used to generate the map of the environment to explore, as the most common sensors, such as RGB-D and stereo cameras, have a limited detection range.
Consequently, UAVs need to fly cautiously to ensure safety.
Cieslewski \emph{et al.} \cite{Cieslewski2017rapid} tackle this limitation, by proposing an exploration strategy that generates velocity commands based on newly detected frontiers, in a bid to maximize the UAV's speed.
This method is shown to outperform classical methods \cite{Yamauchi1997}, but focuses only on local frontiers.
Instead, FUEL \cite{zhou2021fuel} proposes a hierarchical planner which generates efficient global paths, while encouraging safe and agile local maneuvers for high-speed exploration.
FUEL's strategy performs better than \cite{Cieslewski2017rapid} and \cite{Yamauchi1997} in scenes with low obstacle densities.
However, it is more computationally demanding, as it needs to maintain a list of active frontiers, as well as to compute accurate distances between them.
This additional bookkeeping becomes prohibitive and impractical in more cluttered and complex environments such as forests.
In fact, in this type of scenery, the number of frontiers quickly increases due to occlusions caused by tree trunks, branches, and shrubs. 
Another line of research focuses instead on sampling-based path planning to generate viewpoints to explore the space \cite{connolly1985, bircher2016nbv}, by guiding the robot along possible trails of sampled configurations.
The best path is generally found using a greedy approach \cite{bircher2016nbv}, bringing to complete exploration or accurate surface reconstruction \cite{Schmid20ActivePlanning} depending on the information gain formulation.
Nonetheless, these sampled routes may generate trajectories that deviate from the shortest paths, without taking into consideration the robot's dynamics.
Consequently, this causes the UAV to navigate in zigzag patterns, leading to inefficient, slow motions and conservative maneuvers. 
%

%
To tackle the limitations of frontier- and sampling-based methods, also hybrid approaches have been proposed \cite{selin2019efficient, Charrow2015InformationTheoreticPW}.
Such methods compute global paths towards the most informative frontiers while generating local trajectories using sampling-based planners.
However, they do not exploit the full dynamics of the platform and generate sub-optimal routes.
%

%
To boost the efficiency in exploration, various multi-robot cooperative frontier-based methods have also been proposed in the literature, both in centralized \cite{mannucci2018} and decentralized formats \cite{Colares2016TheNF}.
In this spirit, the work in \cite{Hardouin2020} greedily assigns view configurations, while \cite{Dutta2020} distributes the workload between agents using a Voronoi-based partitioning of the area to explore.
Nevertheless, these solutions suffer from the same limitations as in the single-robot case.
Instead, the approach in \cite{morilla2022sweep} is able to generate efficient trajectories for 3D reconstruction, tackling the multi-robot coordination with a centralized architecture.
However, this method requires a prior overhead flight over the area of interest, making it unsuitable for the exploration of forests.
The approach proposed in \cite{yulun2020search} puts more focus on the problem of navigating forests, but the emphasis is more on state estimation rather than on path planning.
Motivated by these limitations, in this work, we propose a strategy that allows a robot to explore complex forest-like environments while flying at high speeds, thanks to the freedom and flexibility that our planner provides to each UAV to switch between different navigation modes online.
While slower, cautious exploration is performed using a frontier-based approach, we efficiently clear trails of unexplored space caused by occlusions by employing a more aggressive local exploration strategy, boosting the efficiency of the mission and pushing the overall time to cover a given area of interest down.
%
%
Moreover, we demonstrate that the proposed pipeline can also be extended to the multi-robot setting in a decentralized fashion.

\section{Methodology}
\label{ch:method}
%
%
\begin{figure*}[!t]%
    \begin{subfigure}{0.4\columnwidth}
      \centering
      \includegraphics[width=\columnwidth]{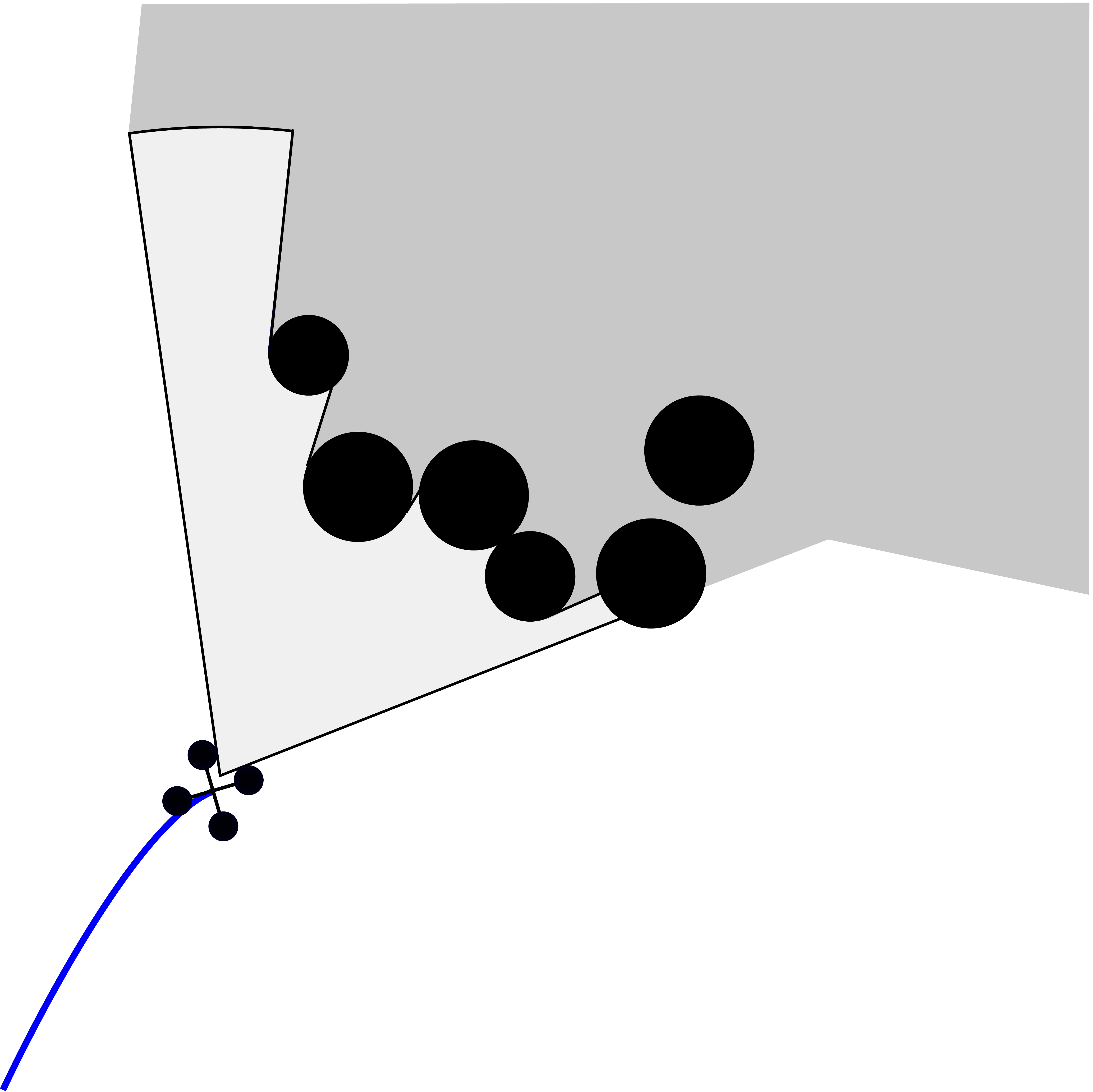}
      \caption{Time-instant 1}
      \label{fig:method:island_1}
    \end{subfigure}%
   \hfil%
   \begin{subfigure}{0.4\columnwidth}
      \centering
      \includegraphics[width=\columnwidth]{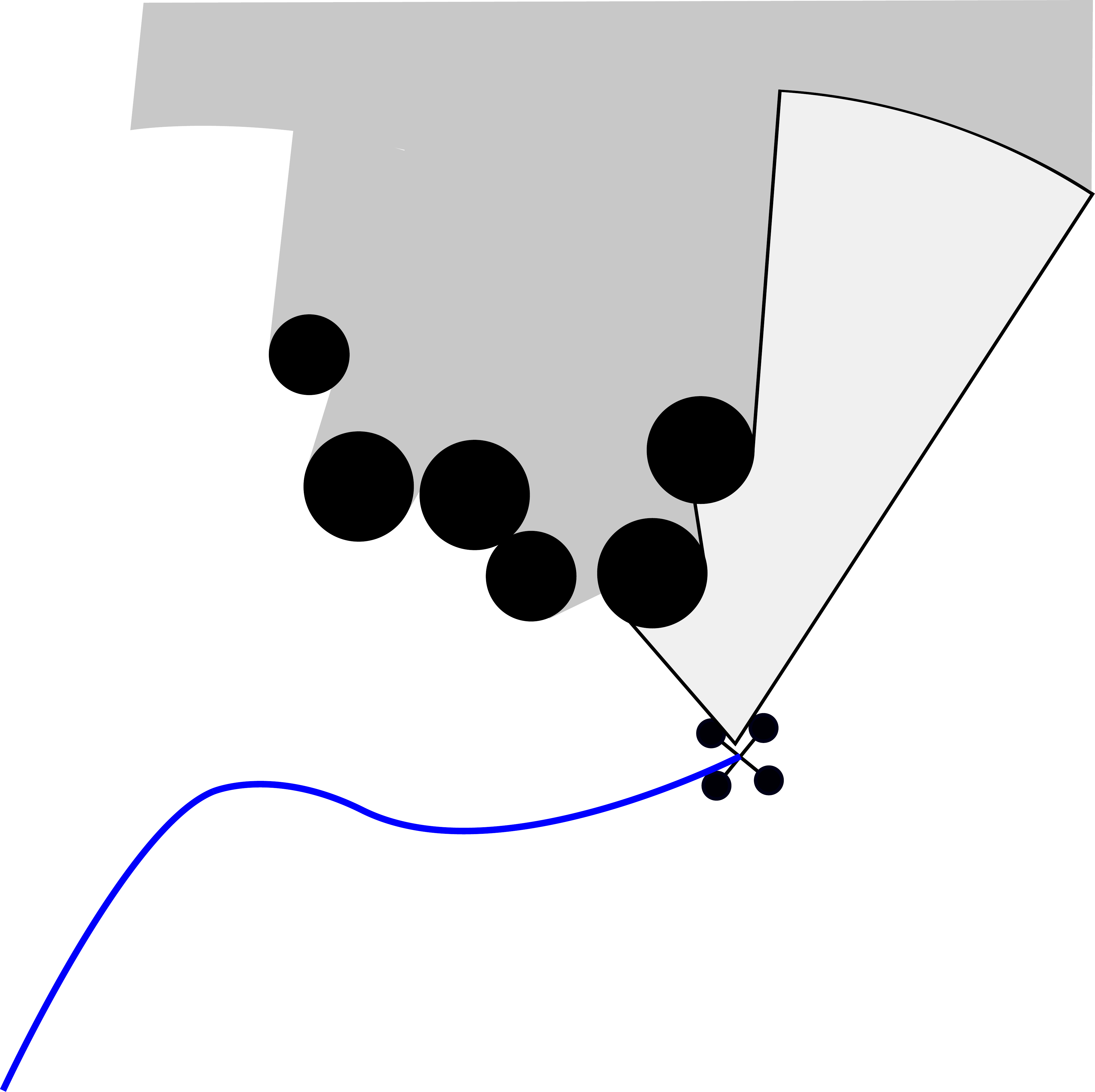}
      \caption{Time-instant 2}
      \label{fig:method:island_2}
    \end{subfigure}%
    \hfil%
    \begin{subfigure}{0.4\columnwidth}
      \centering
      \includegraphics[width=\columnwidth]{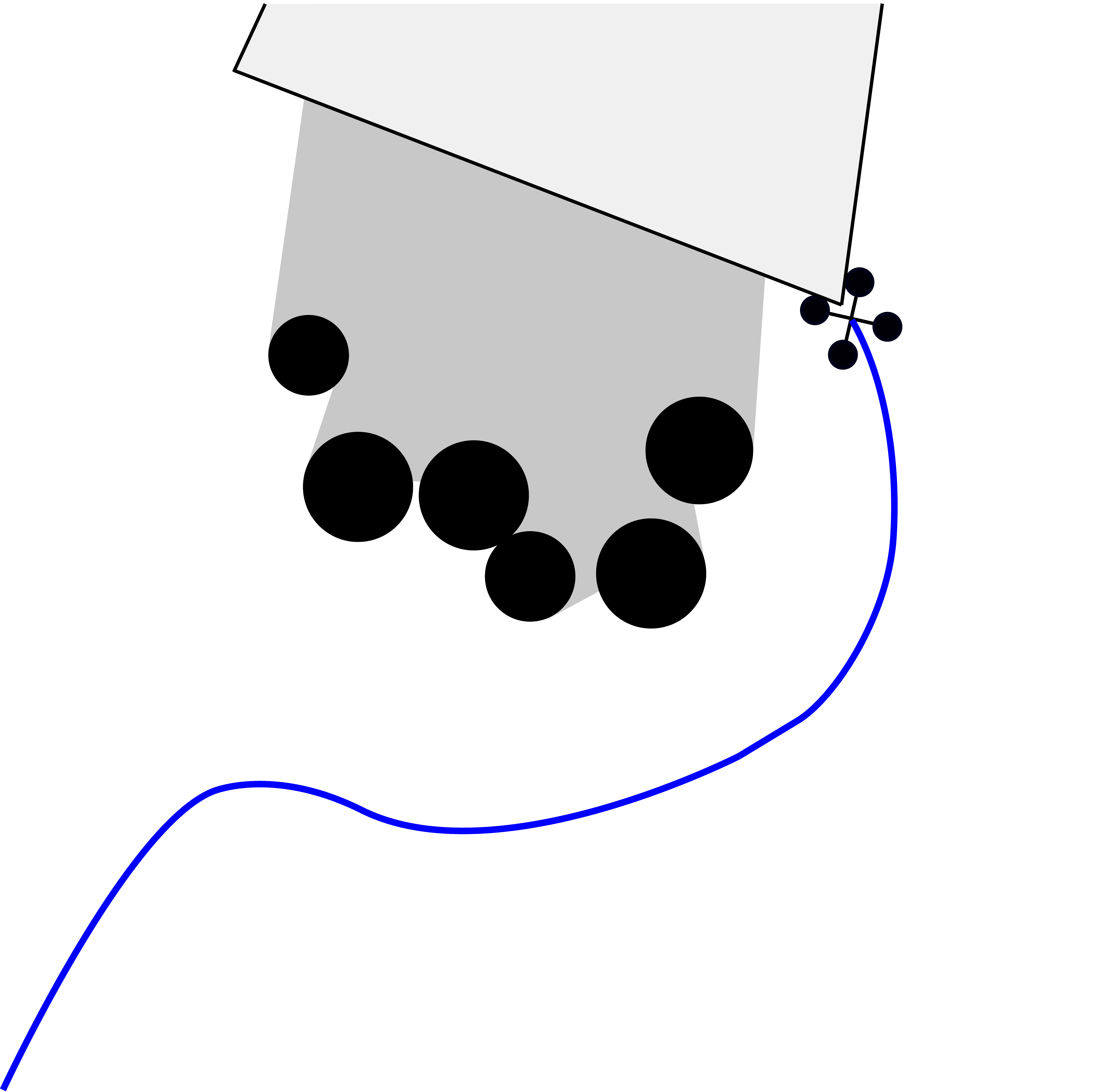}
      \caption{Time-instant 3}
      \label{fig:method:island_3}
    \end{subfigure}%
    \hfil%
    \begin{subfigure}{0.4\columnwidth}
      \centering
      \includegraphics[width=\columnwidth]{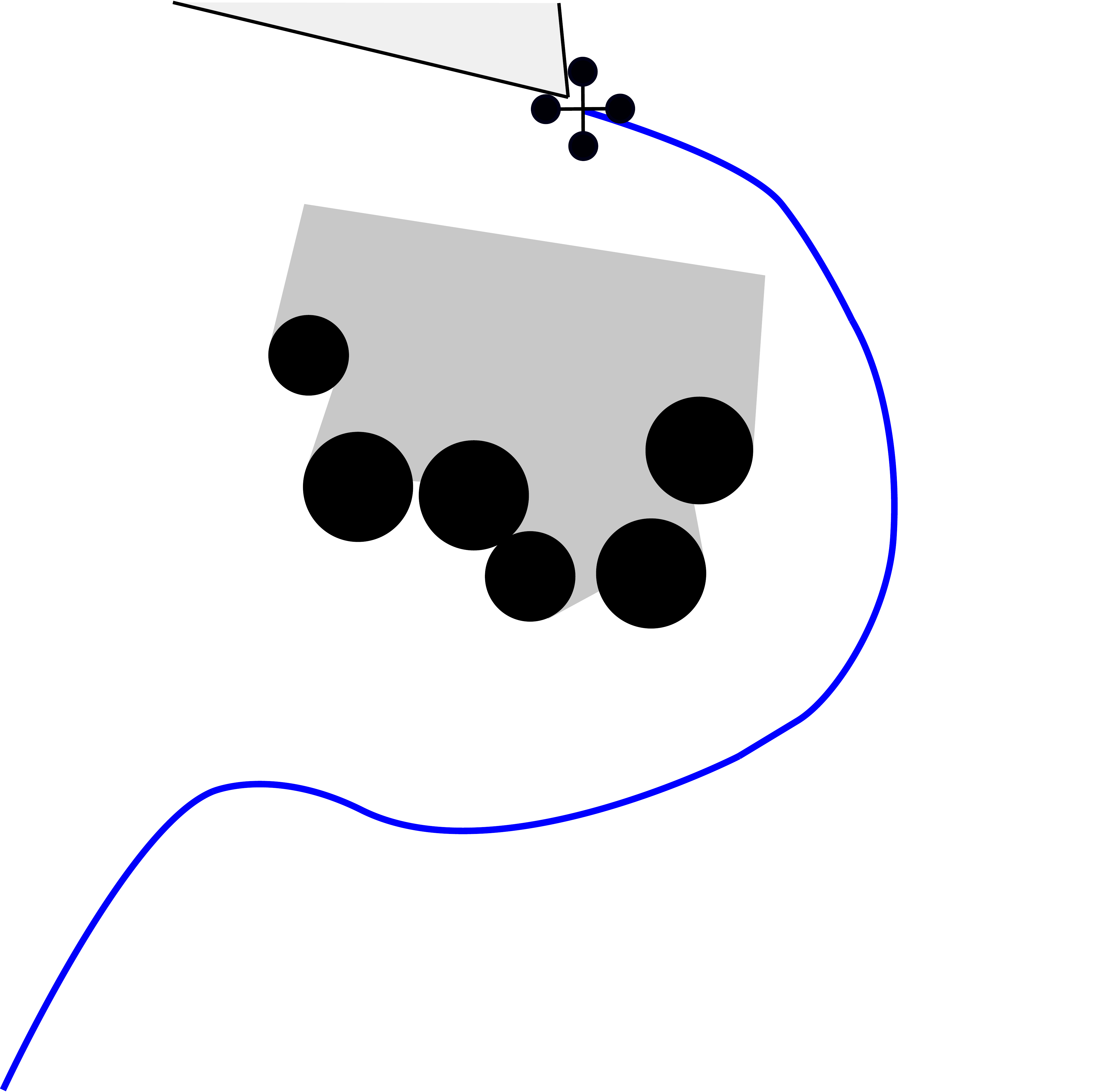}
      \caption{Time-instant 4}
      \label{fig:method:island_4}
    \end{subfigure}%
\caption{
A schematic example demonstrating the problem with greedy frontier-based exploration, at progressive time-instants, generating islands of unknown space surrounded by free regions. The field of view of the robot is depicted as a light-gray shaded area delimited by black solid lines, while the obstacles and the unexplored space are in black and dark gray, respectively (\protect\subref{fig:method:island_1}). The robot navigates towards the most informative frontiers (\protect\subref{fig:method:island_2}); however, due to the limited sensor range, the space occluded by the obstacles is not cleared (\protect\subref{fig:method:island_3}). Consequently, since the exploration process is biased towards larger, more informative frontiers, the na\"ive planner flies the UAV robot ignoring the smaller portion of unexplored space (\protect\subref{fig:method:island_4}).\vspace{-4mm}
}
\label{fig:method:islands}
\end{figure*}
The overall problem considered in this work is to explore unknown cluttered environments, such as forests, in the minimum time possible.
We assume that the robot is equipped with a front-looking depth camera with a limited sensing range and that the robot's odometry information is available at a constant rate.
However, forest-like scenes are characterized by a high number of obstacles in a variety of dimensions (e.g. trunks, leaves, and branches) that make standard frontier-based exploration approaches inefficient.
In fact, during the exploration process, many islands of unknown space are usually left behind, as illustrated in Fig. \ref{fig:method:islands}, necessitating subsequent passes of exploration on a nearly completely explored map.
To tackle this limitation, we propose an exploration pipeline that can change the exploratory behavior of the robot depending on the frontiers in its vicinity. 
In particular, we propose to define two different modes of operation for the robot, namely the \textit{Explorer} and the \textit{Collector} modes.
In the \textit{Explorer} state, the robot is driven by frontiers and it is tasked to explore large unknown areas.
Consequently, it predominantly operates on the most external boundaries between known and unknown areas.
Conversely, the robot in the \textit{Collector} mode clears small islands of unknown space generated by occlusions, that are left behind during the exploratory phase.
The objective of a Collector is to clear these portions of space on the go, avoiding the need for subsequent revisits of the map, at the expense of short local detours.
However, notice that these can be performed at high speed, since, when in Collector mode, the robot operates in mostly explored areas.
By allowing a robot to switch between these two different modes and by finding the right trade-off between map exploration and exploitation, we can quickly reach full coverage of large cluttered environments.
In the following, we first give an overview of the proposed system and our exploration strategies, and then we illustrate how it can be extended to multiple robots.

%

\subsection{System Overview}
\label{ch:method:sys_overview}  
%
\begin{figure*}[t]
	\centering
	\includegraphics[width=\textwidth]{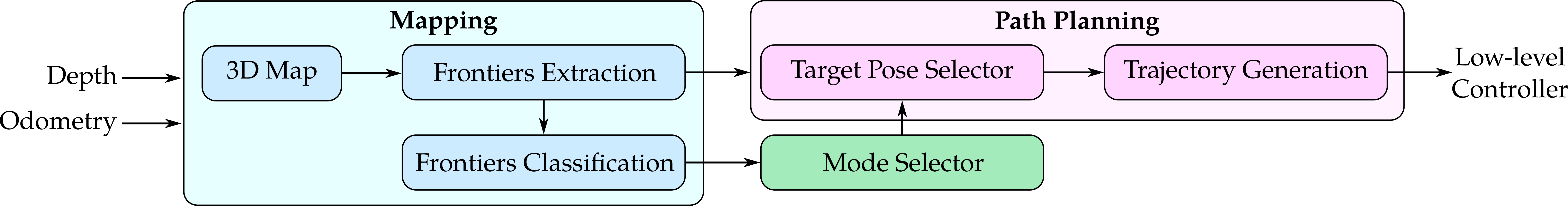}
	\caption{Schematic overview of the proposed exploration pipeline for a single agent. The inputs to the system are the robot's odometry and depth information. These are used to generate a 3D grid-based map of the environment, from which frontiers are extracted and clustered. The trails of frontiers are identified and used to select the adequate exploration mode for the agent. Then, the next target pose is chosen and a trajectory towards the goal pose is generated using \cite{Zhou2019robust}.}
	\label{fig:method:single}
\end{figure*}
As shown in Fig.~\ref{fig:method:single}, the pipeline is composed of three main components: a mapping system, a mode selector, and a path planner.
Given input depth and odometry information, a voxel grid map $\mathcal{M}$ of the environment is generated.
At every update, frontiers are extracted from $\mathcal{M}$ and clustered.
For each cluster, we adopt the sampling strategy from \cite{zhou2021fuel} to generate viewpoints covering the frontiers, and we use them as possible target poses during the exploration process.

Moreover, each cluster undergoes a binary classification step, where unconnected islands of frontiers, or \textit{trails}, are identified.
This is necessary to identify those regions that are likely to require an additional revisiting phase towards the end of the mission if a traditional frontier-based exploration method is utilized.
Here, a cluster is considered a \textit{trail} if its convex hull is surrounded by free space, or when it has only another neighboring cluster.
This implies that most clusters at the corners of the area to be explored are classified as trails.
%
We motivate this design choice by arguing that corners are generally problematic for exploration due to their low informative value.
In fact, they are rarely covered in a first sweep of the map, implying the need for a revisiting step.
%
%
%
%
The labeled clusters are then utilized by the Mode Selector to choose the best exploration strategy for the robot, deciding whether it has to persevere in its current mode, or transit to Explorer or Collector.
The mode assignment is regulated according to the frontiers in the vicinity of the UAV.
Given that our objective is to clear trails locally to avoid large detours on the map, we assign the role of Collector if a minimum number of trails is close to the robot.
Instead, we adopt a more exploratory strategy once all smaller islands of unknown space are cleared, or when the trails are far away from the drone.
Once a strategy is selected, the viewpoint of the most promising cluster is selected as the new target pose.
This is fed to a path planner \cite{Zhou2019robust} that generates the trajectory flying the UAV toward its destination.
We now describe the different exploration modes in more detail.
%

\subsection{Exploration Strategies}
\label{ch:method:strategies}  

\subsubsection{Explorer}
%
Driven by frontiers, the objective of an Explorer is to cover large areas of previously unknown space.
Similarly to \cite{Cieslewski2017rapid}, we process the incoming clusters of frontiers $\mathcal{C}$ from the most recent map update and extract the one with the lowest cost $J_{E}$. 
Notice that these clusters are mostly aligned with the direction of the UAV's motion, implying that, if one of these is selected as the target, the robot avoids abrupt changes in the flight direction or aggressive maneuvers.
The cost associated to the viewpoint $\xi_c := \{\mathbf{x}_c, \gamma_c\}$ covering cluster $c \in \mathcal{C}$ is defined as 
\begin{equation}
    \label{eq:j_e_single}
    J_{E}(\xi_c) := \omega_D J_D(\xi_c) + \omega_V J_V(\xi_c) + \omega_L J_L(c),
\end{equation}
where $\mathbf{x}_c \in \mathbb{R}^3$ is the position of the viewpoint and $\gamma_c \in \mathbb{R}$ its orientation.
The cost $J_D$ is the length of the path in $\mathcal{M}$ between the current robot's position and $\mathbf{x}_c$, and it is calculated using the A* algorithm.
Instead, $J_V$ is associated with the change in direction of travel, while $J_L$ to the label of cluster $c$.
The terms $\omega_D, \omega_V$ and $\omega_L$ are constant weights.
The cost $J_V(\xi_c)$ is calculated as 
\begin{equation}
    J_V(\xi_c) := acos( \mathbf{v}_R^T \frac{\mathbf{x}_c - \mathbf{x}_R}{|| \mathbf{x}_c - \mathbf{x}_R ||_2} ),
\end{equation}
where $\mathbf{v}_R \in \mathbb{R}^3$ and $\mathbf{x}_R \in \mathbb{R}^3$ are the robot's current velocity and position, respectively.
%
This cost is directly associated with the angle between the velocity and the direction vector towards the candidate position $\mathbf{x}_c$ covering cluster $c$.
However, it may happen that the cluster is labeled as \textit{trails}, e.g. in the case of occlusions caused by thin obstacles, such as tree trunks.
Since an Explorer should focus on actual frontiers, we assign a penalty to these clusters:
\begin{equation}
    J_L(c) = 
    \begin{cases}
    0 & \text{if $c$ is}\; frontier \\
    p_{trail} & \text{if $c$ is}\; trail \\
    \end{cases},
\end{equation}
where $p_{trail}$ is the constant penalty associated with trails.
We then select as target pose the next best viewpoint $\xi_{c^*} := \{\mathbf{x}_{c^*}, \gamma_{c^*}\}$ covering the cluster $c^*$ with the lowest cost:
\begin{equation}
    \xi_{c^*} := \arg \min_{\xi_c \; \forall c \in \mathcal{C}}\; J_E(\xi_c).
\end{equation}
In case the UAV is trapped in a dead-end, or if no new clusters are available in front of the robot, we ignore the cost associated with the robot's velocity and we employ a greedy approach to select the new target pose.
We find the best cluster in the vicinity of the robot at a maximum distance $d_{max}$ using the same cost function as in Eq. \ref{eq:j_e_single}, with $J_V$ set to zero:
\begin{align}
    \begin{split}
        \label{eq:j_e_greedy}
        \xi_{c^*} := \arg \min_{\xi_c \; \forall c \in \mathcal{C}} \;\omega_D J_D(\xi_c) + \omega_L J_L(c) \\
        s.t.\; || \mathbf{x}_c - \mathbf{x}_R ||_2 \leq d_{max}.
    \end{split}
\end{align}

\subsubsection{Collector}
%
The objective of a Collector is to clear as many trails as possible, in order to avoid the need for a revisiting step in poorly explored regions of the map at the end of the mission.
Since this task implies a detour from the main direction of exploration, the UAV's speed needs to be maximized in order to go back to Explorer mode as soon as possible.
To reach this objective, we sort the set of trails $\mathcal{C}_{trails} \subseteq \mathcal{C}$ by associating a cost $J_C$ to each cluster $c \in \mathcal{C}_{trails}$:
\begin{equation}
    \label{eq:j_c_single}
    J_C(\xi_c) := \omega_{P} J_P(\xi_c) + \omega_A J_A(\xi_c) ,
\end{equation}
where $J_P$ is associated with the time to reach $\mathbf{x}_c$ and $J_A$ with the time to cover the angular change between the current robot's yaw and the viewpoint's orientation.
Instead, $\omega_P$ and $\omega_A$ are constant weights.
%

%
Given the path $\pi_c^R$ from $\mathbf{x}_R$ to $\mathbf{x}_c$ and the maximum allowed velocity $v_{max}$, $J_P$ is computed as
\begin{equation}
    J_P(\xi_c) := \frac{\text{length}(\pi_c^R)}{v_{max}}.
\end{equation}
Similarly, given the robot current's heading $\gamma_R$, the viewpoint's orientation $\gamma_c$ and the maximum allowed yaw rate $\dot{\gamma}_{max}$, $J_A$ is computed as
\begin{equation}
    J_A(\xi_c) := \frac{\angle(\gamma_R, \gamma_c)}{\dot{\gamma}_{max}},
\end{equation}
where $\angle (\gamma_R, \gamma_c)$ indicates the angular difference between $\gamma_R$ and $\gamma_c$.
The robot then selects the target trail in a step-by-step greedy procedure and behaves as a Collector until all close-by trails are cleared.
%
%
Since the trails are surrounded by free known space, we double the maximum velocity compared to when in Explorer mode.
Consequently, the UAV is able to maximize its velocity, leading to fast motions that allow it to quickly cover all the viewpoints associated with the trails.


\subsection{Extension to Multi-Robot}
\label{ch:method:multi}  
%
\begin{figure}[t]
	\centering
	\includegraphics[width=0.42\textwidth]{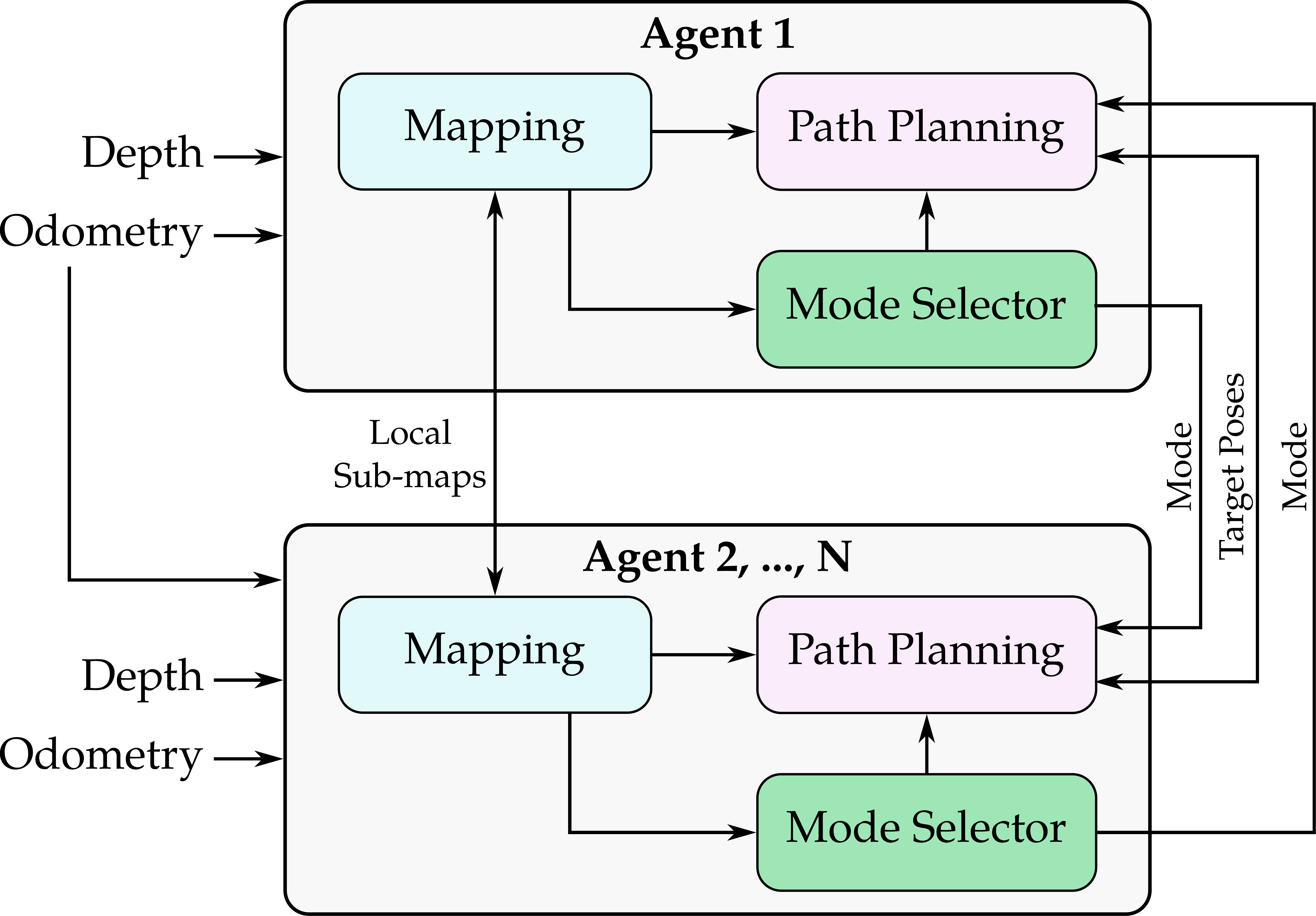}
	\caption{Overview of the pipeline in the multi-robot setting, where the UAVs are tasked to collaboratively build a complete map of the area of interest. To fulfill the objective, agents exchange odometry information and local sub-maps, as well as current execution mode and target poses. We assume there exists a maximum communication range between robots. If the distance between two agents exceeds this limit, communication is lost and information is not exchanged anymore, leading to poor coordination.\vspace{-6mm}}
	\label{fig:method:multi}
\end{figure}
\subsubsection{System Architecture}
The proposed exploration strategy can be easily extended to the multi-robot case. 
The extended pipeline is shown in Fig. \ref{fig:method:multi}.
Assuming that the agents can localize in a common reference frame, they exchange local sub-maps, as well as odometry information, current target pose, and execution mode.
Notice that here we propose a decentralized architecture. 
%
Centralized approaches generally assume infinite-range communication between agents and with a ground station.
However, standard Wi-Fi communications have a limited range and, when navigating in cluttered environments such as forests, communication lines can be potentially obstructed and signal can be lost.
In this work, we propose to use a more flexible point-to-point strategy, assuming that there exists a maximum range of communication between each pair of agents.
%
%
%
%
Our design targets to keep the agents at a valid communication distance.
If the distance between agents is higher than the maximum range, communication is lost and information is not exchanged anymore, leading to poor inter-agent coordination and sub-optimal decision-making.
We also assume that, when communication is lost and successively regained, agents can synchronize their maps.
\subsubsection{Multi-robot Coordination}
%
We encourage coordination between pairs of agents only if they perform compatible actions from an exploration point of view.
This implies that, if two UAVs are operating in the same mode, they can collaborate in order to either explore more unknown spaces (Explorers) or clear trails (Collectors).
On the contrary, coordination between an Explorer and a Collector should not be encouraged, since their tasks are intrinsically different.
In this situation, we propose a \textit{leader-follower} paradigm, where the Explorer (\textit{leader}) explores unknown areas regardless of the position of the second agent, while the Collector (\textit{follower}) follows the leader and clears the trails left unexplored.
This design choice allows more flexibility during the execution of a mission, thanks to the possibility to change execution modes online. 
Notice that, if communication between all agents is lost, their exploration strategy falls back to the single-robot case.
Our collaboration strategy within a robotic team is encouraged as a soft constraint by modifying the cost functions in Eq. \ref{eq:j_e_single} and Eq. \ref{eq:j_c_single}.
If we consider robot $i$ with position $\mathbf{x}_R^i$, given the positions $\mathcal{X}_R^i := \{\mathbf{x}_R^k\}_{k=0}^{N-1}$ of the other $N-1$ robots in the team with $k \neq i$, and their current target positions $\mathcal{G}_R = \{\mathbf{x}_{c^*}^k\}_{k=0}^{N-1}$, we modify the cost functions $J_E$ and $J_C$ for robot $i$ as follows:
\begin{equation}
    \begin{split}
     J_{E}(\xi_c, \mathcal{X}_R^i, \mathcal{G}_R) := \omega_D J_D(\xi_c) + \omega_V J_V(\xi_c) + \\
     \omega_L J_L(c) + \omega_F J_F(\xi_c, \mathcal{X}_R^i, \mathcal{G}_R)
    \end{split}
\end{equation}
and 
\begin{equation}
    \begin{split}
        J_{C}(\xi_c, \mathcal{X}_R^i, \mathcal{G}_R) := \omega_P J_P(\xi_c) + \omega_A J_A(\xi_c) +  \\
        \omega_F J_F(\xi_c, \mathcal{X}_R^i, \mathcal{G}_R),
    \end{split}
\end{equation}
where $\omega_F$ is a constant weight.
The cost function $J_F(\xi_c, \mathcal{X}_R^i, \mathcal{G}_R)$ is defined as 
\begin{equation}
    J_F(\xi_c, \mathcal{X}_R^i, \mathcal{G}_R) := J_F^{att}(\mathcal{X}_R^i) + J_F^{rep}(\xi_c, \mathcal{X}_R^i, \mathcal{G}_R),
\end{equation}
where $J_F^{att}$ aims at keeping the agents $i$ and $k$ in communication range, while $J_F^{rep}$ ensures a minimum distance between them to avoid collisions.
Moreover, $J_F^{rep}$ encourages map splitting, by assigning a high cost to candidate target positions close to other agents' current goals.
In more details, the function $J_F^{att}$ for agent $i$ is defined as follows:
\begin{equation}
    J_F^{att}(\mathcal{X}_R^i) := \sum_{k=0, k \neq i}^{N-1} \mathcal{I}(i,k) \cdot \frac{1}{2} k_A || \mathbf{x}_R^i - \mathbf{x}_R^k ||_2,
\end{equation} where $k_A$ is constant factor and $\mathcal{I}(i, k)$ is an indicator function that embeds our coordination strategy:
\begin{equation}
    \mathcal{I}(i,k) := 
    \begin{cases}
    0 & \text{if}\; i\; \text{Explorer and}\; k\; \text{Collector} \\
    1 & \text{otherwise}
    \end{cases}.
\end{equation}
This indicates that agent $i$ is attracted toward agent $k$ only if they are in a compatible execution mode.
On the contrary, in \textit{leader-follower} mode, i.e. when robot $i$ is an Explorer and robot $k$ is a Collector, the leader ignores the follower, and $J_F^{att}$ goes to zero.
Notice that instead, if $i$ is Collector and $k$ an Explorer, $\mathcal{I}(i,k) = 1$ and $J_F^{att} \neq 0$.
Instead, $J_F^{rep}$ is computed as follows:
\begin{equation}
    J_F^{rep}(\xi_c, \mathcal{X}_R^i, \mathcal{G}_R) := \sum_{k=0, k \neq i}^{N-1} J^{rep}_{ik}(\mathbf{x}_R^i, \mathbf{x}_R^k) + J^{rep}_{ik}(\mathbf{x}_c^i, \mathbf{x}_{c^*}^k),
\end{equation}
where, given the Euclidean distance $d_{AB} := || \mathbf{x}_A - \mathbf{x}_B ||_2$,
\begin{equation}
    J^{rep}_{AB}(\mathbf{x}_A, \mathbf{x}_B) :=
    \begin{cases}
        k_R (d_c - d_0)^2 \frac{d_c d_0}{d_0 - d_c} & \text{if}\; d_{AB} \leq d_0 \\
        k_R (d_{AB} - d_0)^2 & \text{if}\; d_c \leq d_{AB} \leq d_0 \\
        0 & \text{otherwise}
    \end{cases}.
\end{equation}
The parameter $k_R$ is a constant weight, while $d_0$ represents the minimum distance between positions $A$ and $B$ to have a collision.
The parameter $d_c$ represents the distance after which the positions should not approach any closer. 
This can be selected on the basis of the safety distance required between the UAVs.
%
%
Notice that $J^{rep}_{ik}$ is not influenced by the roles of agents $i$ and $k$, as safety and minimum distance requirements need to be always met.


\section{Experiments}
\label{ch:experiments}
We evaluate the proposed exploration pipeline in both single- and multi-UAV setups in simulation.
In particular, we benchmark our method on a series of realistic, randomly generated forests of increasing tree densities \cite{oleynikova2016continuous}, as well as on a 3D reconstruction of a real forest \cite{afzal2021swarm}. 
In the single-agent setup, we compare the proposed method against FUEL \cite{zhou2021fuel}, while in the experiments with multiple robots we test against a 
centralized strategy based on map-splitting.
%
In all tests, we use grid map resolutions of $\SI{0.10}{\meter}$ or $\SI{0.15}{\meter}$ depending on the map size, while we set the dynamic limits to $v_{max} = \SI{1.5}{\meter / \second}$ and $\dot{\gamma}_{max} = \SI{0.9}{rad / \second}$ for all planners.
%
%
We simulate a depth camera with a fixed range of $\SI{4.5}{\meter}$ using the Vulkan-based renderer of \cite{bartolomei2022multi} and the same physical simulator as in \cite{zhou2021fuel}. 
We report the planners' performance in terms of the time needed to complete the exploration of the scene, the total travelled distance, and the average velocity of the UAVs during each experiment.
%
%

\subsection{Single-robot Experiments}
\begin{table}[t!]
	\renewcommand{\arraystretch}{1.2}
	\begin{center}
		\resizebox{0.45\textwidth}{!}{
		\begin{tabular}{l|ccc}
			\toprule
			& Ours & FUEL \cite{zhou2021fuel}\\
			\toprule

			\multicolumn{3}{c}{\textsc{Real Forest}} \\
   
			Completion Time [$\SI{}{\second}$] & \bf{500.7 $\pm$ 14.8} & 757.7 $\pm$ 47.9  \\
			Travelled Distance [$\SI{}{\meter}$] & 645.0 $\pm$ 20.0 & \bf{533.2 $\pm$ 11.0}  \\
			Velocity [$\SI{}{\meter / \second}$] & \bf{1.3 $\pm$ 0.5} & 0.7 $\pm$ 0.4  \\
			\midrule

			\multicolumn{3}{c}{\textsc{Sparse Forest (0.05 trees / }$\SI{}{\meter}^2$)} \\
			Completion Time [$\SI{}{\second}$] & \bf{665.4 $\pm$ 32.7} & 1114.1 $\pm$ 97.4  \\
			Travelled Distance [$\SI{}{\meter}$] & 860.9 $\pm$ 34.0 & \bf{758.1 $\pm$ 57.6}  \\
			Velocity [$\SI{}{\meter / \second}$] & \bf{1.3 $\pm$ 0.6} & 0.7 $\pm$ 0.5  \\
			\midrule

			\multicolumn{3}{c}{\textsc{Average-Density Forest (0.10 trees / }$\SI{}{\meter}^2$)} \\
			Completion Time [$\SI{}{\second}$] & \bf{779.6 $\pm$ 110.9} & 954.1 $\pm$ 28.8 \\
			Travelled Distance [$\SI{}{\meter}$] & 910.3 $\pm$ 63.7 & \bf{713.5 $\pm$ 33.1}  \\
			Velocity [$\SI{}{\meter / \second}$] & \bf{1.2 $\pm$ 0.6} & 0.7 $\pm$ 0.5  \\
			\midrule

			\multicolumn{3}{c}{\textsc{Dense Forest (0.15 trees / }$\SI{}{\meter}^2$)} \\
			Completion Time [$\SI{}{\second}$] & \bf{613.2 $\pm$ 16.2} & 1130.2 $\pm$ 28.8  \\
			Travelled Distance [$\SI{}{\meter}$] & \bf{789.7 $\pm$ 16.8} & 791.0 $\pm$ 37.7  \\
			Velocity [$\SI{}{\meter / \second}$] & \bf{1.2 $\pm$ 0.6} & 0.7 $\pm$ 0.5  \\
			\midrule

			\multicolumn{3}{c}{\textsc{Very Dense Forest (0.20 trees / }$\SI{}{\meter}^2$)} \\
			Completion Time [$\SI{}{\second}$] & \bf{658.2 $\pm$ 57.2} & 904.1 $\pm$ 109.5  \\
			Travelled Distance [$\SI{}{\meter}$] & 802.0 $\pm$ 52.7 & \bf{680.6 $\pm$ 49.9}  \\
			Velocity [$\SI{}{\meter / \second}$] & \bf{1.2 $\pm$ 0.6} & 0.7 $\pm$ 0.5  \\
			\bottomrule
		\end{tabular}
		}
	\end{center}
	\renewcommand{\arraystretch}{1}
	\caption{Results of the experiments for a single agent. We report the average completion time over 3 runs, as well as the average travelled distance and velocity. The best performance is in bold. The relatively high standard deviation in the timings to complete the missions, in particular in the cases of tree densities of $0.10$ and $0.20$ trees/m$^2$, is caused by the complex nature of the map and the large number of occlusions.\vspace{-3mm}}
	\label{table:exp:single}
\end{table}
\begin{figure}[t]
	\centering
	\includegraphics[width=0.45\textwidth]{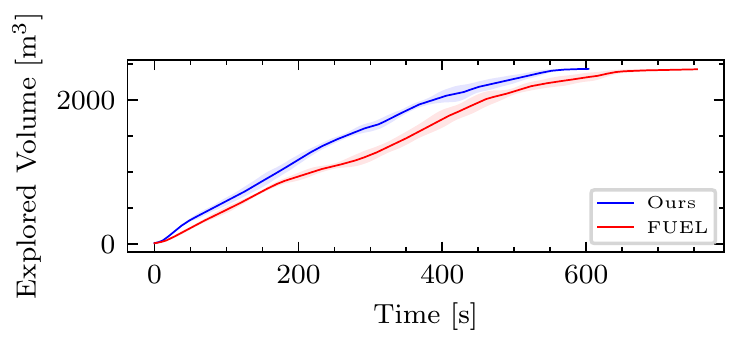} \vspace{-2mm}
	\caption{The average exploration rate during the experiments in the model \textsc{Real Forest}. The shaded region shows the standard deviation. The proposed method reaches complete coverage in less time than FUEL \cite{zhou2021fuel}.\vspace{-3mm}}
	\label{fig:exp:cov_single}
\end{figure}
\begin{figure}[t]
	\centering
	\includegraphics[width=0.45\textwidth]{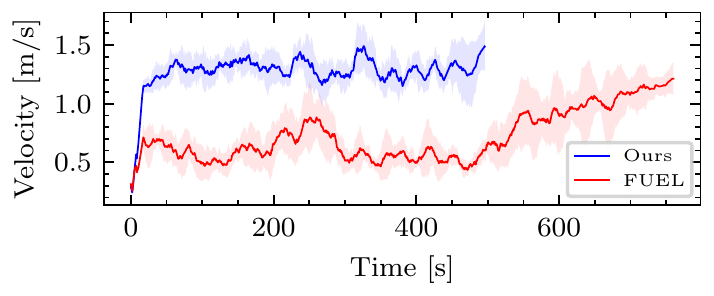} \vspace{-2mm}
	\caption{The average UAV velocity during the experiments in the \textsc{Real Forest}. The shaded region indicates the standard deviation. The proposed strategy is able to fly the UAV at higher velocities than FUEL leading to improved mission efficiency (i.e. time to mission completion). Towards the end of the mission, the UAV following the FUEL strategy also speeds up, as by then the map is mostly explored, and smaller trails are cleared, resembling the Collector mode in the proposed strategy.\vspace{-3mm}}
	\label{fig:exp:vel_single}
\end{figure}
%
%
%
%
In the single-robot experiments, the models of the synthetic forests are of size $\SI{50}{\meter} \times \SI{50}{\meter} \times \SI{2}{\meter}$, while the 3D reconstruction of the \textsc{Real Forest} has dimensions $\SI{40}{\meter} \times \SI{40}{\meter} \times \SI{2}{\meter}$. 
The map resolution is set to $\SI{0.10}{\meter}$.
As reported in Table \ref{table:exp:single}, the proposed planner outperforms FUEL \cite{zhou2021fuel} across all scenes in terms of the time taken to reach full coverage (Fig. \ref{fig:exp:cov_single}), thanks to our adaptive exploration policy that leads to consistently higher UAV velocity throughout each mission, as illustrated in Fig. \ref{fig:exp:vel_single}.
%
%
These results demonstrate the benefit of using the proposed adaptive exploration strategy over a fixed-mode method.
However, notice that the proposed design leads to longer travelled distances, albeit guaranteeing that there are no small unexplored areas left.
In fact, decision-making both in Explorer and Collector modes is done on a local-map level, and this may cause the UAV to fly longer routes, deviating from the shortest path.
Nonetheless, in the proposed strategy we compensate for this shortcoming by encouraging decisions leading to higher UAV velocities, and thus shorter mission times.

%

%
\subsection{Multi-robot Experiments}
%
%
%
In the multi-robot experiments, the proposed planning strategy is tested in a variety of models with fixed, homogeneous obstacle density, as well as in a randomly generated forest with tree density varying across different regions of the map.

\subsubsection{Maps with a fixed tree density}
\label{sec:exp:multi:fixed}
\begin{figure}[t!]
  \centering
  \includegraphics[width=0.95\linewidth]{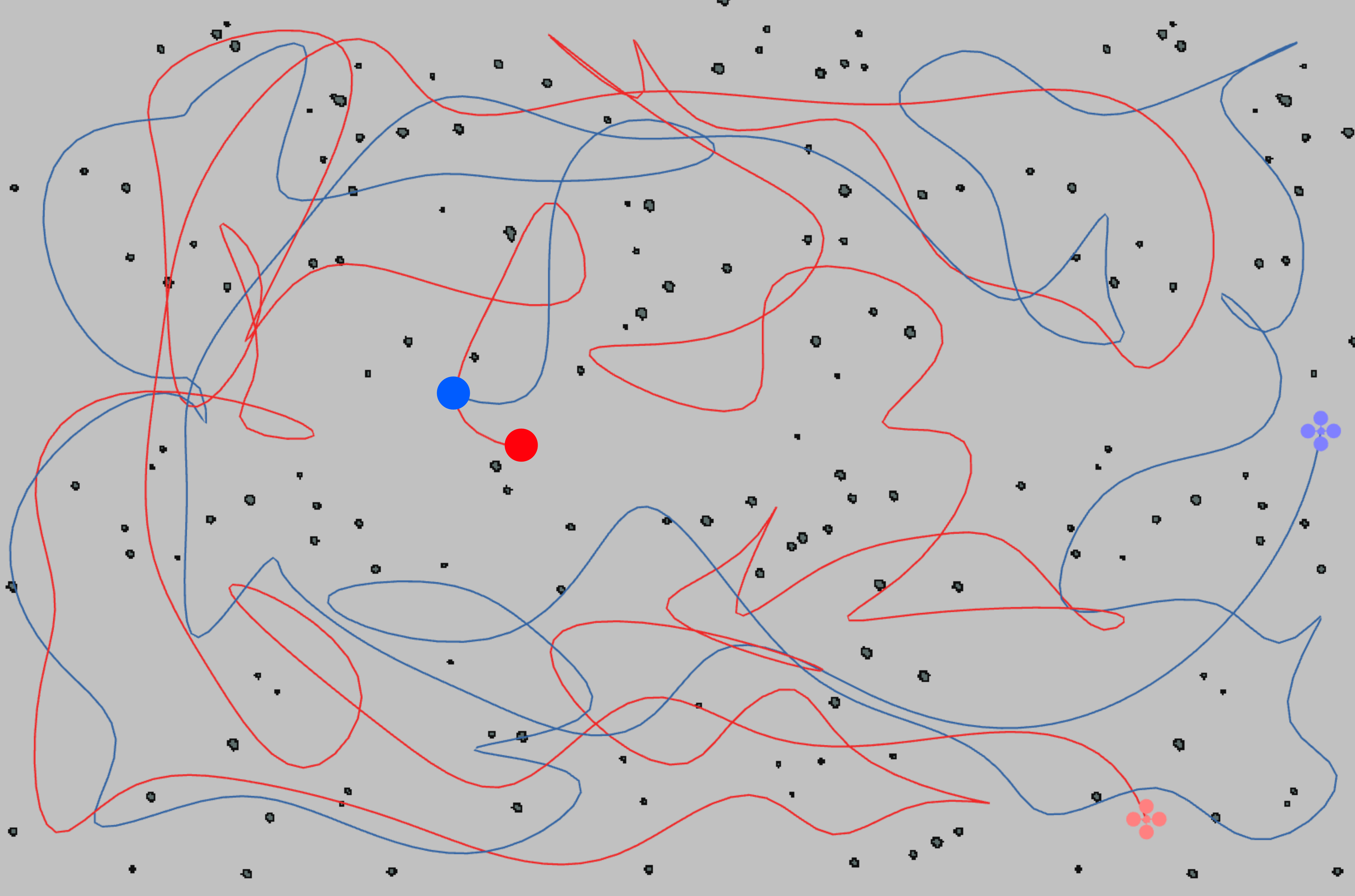}
  \caption{Top view of an exploration mission, where a team of two UAVs is tasked to map a randomly generated forest with tree density $0.05$ trees/m$^2$, illustrated with the black dots here. The initial positions of the UAVs are denoted as colored blobs and the final positions as enlarged drone models for clearer visualization. The map is represented as a 2D occupancy grid obtained by slicing the 3D model at $\SI{1.5}{\meter}$ from the ground.\vspace{-2mm}}
  \label{fig:exp:005_two}
\end{figure}
\begin{table}[t!]
	\renewcommand{\arraystretch}{1.0}
	\begin{center}
		\resizebox{0.475\textwidth}{!}{
		\begin{tabular}{l|cc}
			\toprule
			 & Ours & Split Map (FUEL \cite{zhou2021fuel}) \\
			\toprule

			\multicolumn{3}{c}{\textsc{Sparse Forest (0.05 trees / }$\SI{}{\meter}^2$)} \\
			Completion Time [$\SI{}{\second}$]   & \bf{780.3 $\pm$ 32.2} & 834.3 $\pm$ 64.3 \\
			Travelled Distance [$\SI{}{\meter}$] & 958.8 $\pm$ 9.2       & \bf{595.7 $\pm$ 51.8} \\
                                                     & 958.7 $\pm$ 60.4      & \\
			Velocity [$\SI{}{\meter / \second}$] & \bf{1.2 $\pm$ 0.6}    & 0.7 $\pm$ 0.4 \\
                                                     & \bf{1.3 $\pm$ 0.7}    & \\
			\midrule

			\multicolumn{3}{c}{\textsc{Average-Density Forest (0.10 trees / }$\SI{}{\meter}^2$)} \\
			Completion Time [$\SI{}{\second}$]   & \bf{838.5 $\pm$ 64.4} & 848.1 $\pm$ 98.5 \\
			Travelled Distance [$\SI{}{\meter}$] & 915.3 $\pm$ 68.4      & \bf{616.1 $\pm$ 40.7} \\
                                                     & 906.2 $\pm$ 58.3      & \\
			Velocity [$\SI{}{\meter / \second}$] & \bf{1.2 $\pm$ 0.5}    & 0.8 $\pm$ 0.4 \\
                                                     & \bf{1.1 $\pm$ 0.6}    & \\
			\midrule

			\multicolumn{3}{c}{\textsc{Dense Forest (0.15 trees / }$\SI{}{\meter}^2$)} \\
			Completion Time [$\SI{}{\second}$]   & 786.3 $\pm$ 40.7      & \bf{754.0 $\pm$ 23.9} \\
			Travelled Distance [$\SI{}{\meter}$] & 839.0 $\pm$ 26.1      & \bf{586.7 $\pm$ 14.2} \\
                                                     & 882.2 $\pm$ 32.0      &  \\
			Velocity [$\SI{}{\meter / \second}$] & \bf{1.2 $\pm$ 0.7}    & 0.8 $\pm$ 0.4 \\
                                                     & \bf{1.3 $\pm$ 0.7}    & \\
			\midrule

			\multicolumn{3}{c}{\textsc{Very Dense Forest (0.20 trees / }$\SI{}{\meter}^2$)} \\
			Completion Time [$\SI{}{\second}$]   & 803.7 $\pm$ 52.8      & \bf{705.8 $\pm$ 73.2} \\
			Travelled Distance [$\SI{}{\meter}$] & 873.8 $\pm$ 47.0      & \bf{580.3 $\pm$ 73.3} \\
                                                     & 912.6 $\pm$ 42.7      & \\
			Velocity [$\SI{}{\meter / \second}$] & \bf{1.2 $\pm$ 0.7}    &  0.7 $\pm$ 0.5 \\
                                                     & \bf{1.2 $\pm$ 0.8}    &  \\
			\midrule

			\bottomrule
		\end{tabular}
		}
	\end{center}
	\renewcommand{\arraystretch}{1}
	\caption{Results in randomly generated forests with fixed tree densities explored with two UAVs, averaged over 3 runs. For the proposed strategy we report the average travelled distance and velocity per agent. The best performance is highlighted in bold.\vspace{-5mm}}
	\label{table:exp:multi_2}
\end{table}
\begin{figure}[t]
	\centering
	\includegraphics[width=0.45\textwidth]{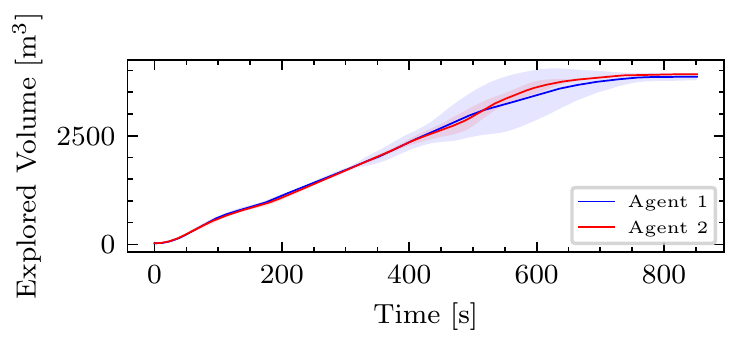}
	\caption{The average exploration rate per agent with the proposed approach using two UAVs during the experiments in a random forest with density $0.10$ trees/$\SI{}{\meter}^2$. The shaded region shows the standard deviation. The explored volume is shown to be consistently balanced across the two agents.\vspace{-5mm}}
	\label{fig:exp:cov_multi}
\end{figure}
The results of the multi-robot collaborative exploration strategy of maps with fixed tree densities with two agents are shown in Table \ref{table:exp:multi_2}, where a maximum connection distance of $\SI{50}{\meter}$ for data exchange between agents is assumed (Fig. \ref{fig:exp:005_two}).
Here, experiments are performed in forest models of size $\SI{100}{\meter} \times \SI{50}{\meter} \times \SI{2}{\meter}$ with a map resolution of $\SI{0.10}{\meter}$.
%
The proposed approach is compared against a centralized strategy we devised, employing the FUEL planner \cite{zhou2021fuel} and assigning the UAVs to explore maps of equal sizes (i.e. using map-splitting). 
Note that this strategy assumes homogeneous forest maps, and knowledge of the original map size, which renders this unsuitable for realistic deployment unlike the proposed approach; however, comparisons are presented for the sake of benchmarking.
%
%

%
The proposed strategy reaches comparable results with respect to the centralized approach using FUEL.
Similarly to single-robot exploration, the proposed strategy flies the UAVs at higher speeds, incurring longer travel distances.
Moreover, our strategy enables automatic load balancing of the exploration mission, yielding similar exploration rates per agent as illustrated in Fig. \ref{fig:exp:cov_multi}.
However, in denser forest models, the performance of the proposed approach is seen to degrade, as the UAVs are tasked to fly within a connection range to each other, resulting in limited freedom of movement.
This is exacerbated by the increased number of obstacles and occlusions in smaller maps, leading to lower exploration rates.
Nevertheless, as aforementioned, the proposed approach is realistically deployable in contrast to the baseline strategy using FUEL.

\subsubsection{Map with non-homogeneous tree density}
The results of the experiments in a map with non-homogeneous obstacle densities with a team of two agents are shown in Table \ref{table:exp:multi_densities}.
We utilize a model with size $\SI{100}{\meter} \times \SI{200}{\meter} \times \SI{2}{\meter}$, with different tree densities ($0.2$, $0.3$ and $0.5$ trees/$\SI{}{\meter}^2$) across distinct map regions.
The occupancy grip map resolution is set to $\SI{0.15}{\meter}$, with a maximum inter-agent communication range of $\SI{200}{\meter}$.
We perform the same analysis as in Sec. \ref{sec:exp:multi:fixed}, assuming a realistic maximum flight time of $\SI{1500}{\second}$ for the UAVs.
As none of the tested planners is able to fully explore the environment within the allowed time, we report the total team coverage at fixed timestamps.
%
%
The proposed approach consistently outperforms the solution based on map-splitting using FUEL \cite{zhou2021fuel} as the planning back-end.
The gain in performance is related to the capacity of our strategy to fly the UAVs faster in regions with lower obstacle density and to explore cautiously more cluttered areas while clearing smaller frontier trails on the go.
This leads to a more efficient exploration process, able to better exploit the capacity of UAVs to perform highly dynamic flights.
Finally, we report the performance of our method when a team of three robots is deployed.
As shown in Table \ref{table:exp:team_size}, a larger team size yields higher total coverage.
%
%
This demonstrates that this work presents an effective strategy for peer-to-peer sharing of the responsibility of exploration of a large forest area and that the extension to larger teams of multiple UAVs can be realized following this paradigm, pushing for scalable, decentralized multi-robot planning for exploration.
\begin{table}[t!]
	\renewcommand{\arraystretch}{1.0}
	\begin{center}
		\resizebox{0.475\textwidth}{!}{
		\begin{tabular}{l|cc}
			\toprule
			 & Ours & Split Map (FUEL \cite{zhou2021fuel}) \\
			\toprule
			Travelled Distance [$\SI{}{\meter}$] & 1481.9 $\pm$ 467.0 & \bf{735.2 $\pm$ 500.1} \\
                                                     & 1487.0 $\pm$ 440.8 & \bf{697.1 $\pm$ 365.6} \\
            & & \\
			Velocity [$\SI{}{\meter / \second}$] & \bf{1.3 $\pm$ 0.6} & 0.7 $\pm$ 0.4 \\
                                                     & \bf{1.2 $\pm$ 0.7} & 0.6 $\pm$ 0.5 \\
            & & \\
            %
		    Explored Volume [$\SI{}{\meter}^3$] - $\SI{300}{\second}$ &     \bf{542.3 $\pm$ 23.3}       &    514.6 $\pm$ 313.0  \\
                                                                          &     565.9 $\pm$ 31.4            &    \bf{632.0 $\pm$ 184.8} \\
            & & \\
			Explored Volume [$\SI{}{\meter}^3$] - $\SI{600}{\second}$ &     \bf{1062.8 $\pm$ 81.3}      &    840.0 $\pm$ 580.7 \\
                                                                          &     \bf{1011.6 $\pm$ 65.3}      &    728.3 $\pm$ 163.5     \\
            & & \\
			Explored Volume [$\SI{}{\meter}^3$] - $\SI{900}{\second}$ &     \bf{1400.0 $\pm$ 265.3}     &    1145.3 $\pm$ 814.7 \\
                                                                          &     \bf{1302.6 $\pm$ 110.5}     &    798.8 $\pm$ 146.2     \\
            & & \\
			Explored Volume [$\SI{}{\meter}^3$] - $\SI{1200}{\second}$ &     \bf{1626.7 $\pm$ 431.2}     &    1330.1 $\pm$ 905.3 \\
                                                                           &     \bf{1466.5 $\pm$ 185.4}     &    910.4 $\pm$ 83.9       \\
            & & \\
			Explored Volume [$\SI{}{\meter}^3$] - $\SI{1500}{\second}$ &     \bf{1816.0 $\pm$ 550.8}     &    1437.8 $\pm$ 971.8 \\
                                                                           &     \bf{1637.0 $\pm$ 299.1}     &    1040.5 $\pm$ 124.3 \\
			\bottomrule
		\end{tabular}
		}
	\end{center}
	\renewcommand{\arraystretch}{1}
	\caption{Results in a randomly generated forest with non-homogeneous tree densities when explored with two UAVs, averaged over 3 runs. We report the average travelled distance and velocity per agent at $\SI{1500}{\second}$, as well as the total explored volume by the team at different timestamps. The best performance is highlighted in bold.}
	\label{table:exp:multi_densities}
\end{table}
\begin{table}[t!]
	\renewcommand{\arraystretch}{1.2}
	\begin{center}
		\resizebox{0.4\textwidth}{!}{
		\begin{tabular}{c|cc}
			\toprule
        Timestamp & Two Agents & Three Agents \\
			\toprule
        $\SI{300}{\second}$ & $1108.2 \pm \SI{54.7}{\meter^3}$ & \bf{1208.0 $\pm$ 91.6 m$^3$} \\
        & & \\
        $\SI{600}{\second}$ & $2074.3 \pm \SI{144.8}{\meter^3}$ & \bf{2098.4 $\pm$ 143.8 m$^3$} \\
        & & \\
        $\SI{900}{\second}$ & $2702.6 \pm \SI{375.9}{\meter^3}$ & \bf{2748.3 $\pm$ 111.9 m$^3$} \\
        & & \\
        $\SI{1200}{\second}$ & $3093.2 \pm \SI{616.6}{\meter^3}$ & \bf{3223.5 $\pm$ 127.9 m$^3$} \\
        & & \\
        $\SI{1500}{\second}$ & $3453.0 \pm \SI{849.9}{\meter^3}$ & \bf{3621.8 $\pm$ 321.0 m$^3$} \\
			\bottomrule
		\end{tabular}
		}
	\end{center}
	\renewcommand{\arraystretch}{1}
	\caption{Results in a randomly generated forest with varying tree densities across different regions of the model, averaged over 3 runs. Here, the map is explored with teams composed of two and three UAVs. We report the total volume covered by the team at different timestamps. The best performance is highlighted in bold.\vspace{-3mm}}
	\label{table:exp:team_size}
\end{table}
\section{Conclusion and Future Work}
\label{ch:conclusions}
In this work, we propose an exploration pipeline for autonomous UAVs operating in complex, cluttered environments, with a particular focus on forests. 
We choose this type of environment as one of the inherently most challenging for effective planning due to the increased number of obstacles and occlusions that they exhibit.
The proposed strategy allows each UAV to switch between different exploratory behaviors, autonomously balancing cautious exploration of unknown space and more aggressive maneuvers, exploiting already mapped space within a mission.
This leads to faster completion times due to higher-speed flights and, consequently, to more efficient and faster map coverage than the state of the art.
%
Moreover, we show how the proposed method can be extended to three, and potentially more robots in a decentralized fashion, demonstrating automatic and effective load balancing across the participating agents.
Following the push for automating higher-level decision-making in robotic missions, this work constitutes a key milestone towards effective exploration planning for robotic teams.
The natural next step for this work is to address the integration and deployment of the proposed pipeline onboard real platforms, while further investigations will push advancing coordination strategies in larger multi-robot teams.

\balance
\bibliographystyle{IEEEtran}
\bibliography{references}

\begin{thebibliography}{10}
\providecommand{\url}[1]{#1}
\csname url@samestyle\endcsname
\providecommand{\newblock}{\relax}
\providecommand{\bibinfo}[2]{#2}
\providecommand{\BIBentrySTDinterwordspacing}{\spaceskip=0pt\relax}
\providecommand{\BIBentryALTinterwordstretchfactor}{4}
\providecommand{\BIBentryALTinterwordspacing}{\spaceskip=\fontdimen2\font plus
\BIBentryALTinterwordstretchfactor\fontdimen3\font minus
  \fontdimen4\font\relax}
\providecommand{\BIBforeignlanguage}[2]{{%
\expandafter\ifx\csname l@#1\endcsname\relax
\typeout{** WARNING: IEEEtran.bst: No hyphenation pattern has been}%
\typeout{** loaded for the language `#1'. Using the pattern for}%
\typeout{** the default language instead.}%
\else
\language=\csname l@#1\endcsname
\fi
#2}}
\providecommand{\BIBdecl}{\relax}
\BIBdecl

\bibitem{selin2019efficient}
M.~Selin, M.~Tiger, D.~Duberg, F.~Heintz, and P.~Jensfelt, ``{Efficient
  Autonomous Exploration Planning of Large-Scale 3-D Environments},''
  \emph{IEEE Robotics and Automation Letters}, 2019.

\bibitem{Schmid20ActivePlanning}
L.~{Schmid}, M.~{Pantic}, R.~{Khanna}, L.~{Ott}, R.~{Siegwart}, and J.~{Nieto},
  ``{An Efficient Sampling-Based Method for Online Informative Path Planning in
  Unknown Environments},'' \emph{IEEE Robotics and Automation Letters}, 2020.

\bibitem{kompis2021informed}
Y.~Kompis, L.~Bartolomei, R.~Mascaro, L.~Teixeira, and M.~Chli, ``{Informed
  Sampling Exploration Path Planner for 3D Reconstruction of Large Scenes},''
  \emph{IEEE Robotics and Automation Letters}, 2021.

\bibitem{Cieslewski2017rapid}
T.~Cieslewski, E.~Kaufmann, and D.~Scaramuzza, ``{Rapid exploration with
  multi-rotors: A frontier selection method for high speed flight},'' in
  \emph{2017 IEEE/RSJ International Conference on Intelligent Robots and
  Systems (IROS)}, 2017.

\bibitem{zhou2021fuel}
B.~Zhou, Y.~Zhang, X.~Chen, and S.~Shen, ``{FUEL: Fast UAV Exploration Using
  Incremental Frontier Structure and Hierarchical Planning},'' \emph{IEEE
  Robotics and Automation Letters}, 2021.

\bibitem{afzal2021swarm}
A.~Ahmad, V.~Walter, P.~Petráček, M.~Petrlík, T.~Báča, D.~Žaitlík, and
  M.~Saska, ``{Autonomous Aerial Swarming in GNSS-denied Environments with High
  Obstacle Density},'' in \emph{2021 IEEE International Conference on Robotics
  and Automation (ICRA)}, 2021.

\bibitem{yulun2020search}
Y.~Tian, K.~Liu, K.~Ok, L.~Tran, D.~Allen, N.~Roy, and J.~P. How, ``{Search and
  rescue under the forest canopy using multiple UAVs},'' \emph{The
  International Journal of Robotics Research}, 2020.

\bibitem{rouvcek2019darpa}
T.~Rou{\v{c}}ek, M.~Pecka, P.~{\v{C}}{\'\i}{\v{z}}ek,
  T.~Pet{\v{r}}{\'\i}{\v{c}}ek, J.~Bayer, V.~{\v{S}}alansk{\`y}, D.~He{\v{r}}t,
  M.~Petrl{\'\i}k, T.~B{\'a}{\v{c}}a, V.~Spurn{\`y} \emph{et~al.}, ``Darpa
  subterranean challenge: Multi-robotic exploration of underground
  environments,'' in \emph{International Conference on Modelling and Simulation
  for Autonomous Systems}.\hskip 1em plus 0.5em minus 0.4em\relax Springer,
  2019.

\bibitem{bartolomei2020server}
L.~Bartolomei, M.~Karrer, and M.~Chli, ``{Multi-robot Coordination with
  Agent-Server Architecture for Autonomous Navigation in Partially Unknown
  Environments},'' in \emph{2020 IEEE/RSJ International Conference on
  Intelligent Robots and Systems (IROS)}, 2020.

\bibitem{corah2019communication}
M.~Corah, C.~O’Meadhra, K.~Goel, and N.~Michael, ``Communication-efficient
  planning and mapping for multi-robot exploration in large environments,''
  \emph{IEEE Robotics and Automation Letters}, 2019.

\bibitem{Yamauchi1997}
B.~Yamauchi, ``A frontier-based approach for autonomous exploration,'' in
  \emph{Proceedings 1997 IEEE International Symposium on Computational
  Intelligence in Robotics and Automation CIRA'97}, 1997.

\bibitem{dasilva2020}
D.~L. da~Silva~Lubanco, M.~Pichler-Scheder, and T.~Schlechter, ``A novel
  frontier-based exploration algorithm for mobile robots,'' in \emph{2020 6th
  International Conference on Mechatronics and Robotics Engineering (ICMRE)},
  2020.

\bibitem{connolly1985}
C.~Connolly, ``The determination of next best views,'' in \emph{Proceedings.
  IEEE International Conference on Robotics and Automation}, 1985.

\bibitem{bircher2016nbv}
A.~Bircher, M.~Kamel, K.~Alexis, H.~Oleynikova, and R.~Siegwart, ``{Receding
  Horizon "Next-Best-View" Planner for 3D Exploration},'' in \emph{2016 IEEE
  International Conference on Robotics and Automation (ICRA)}, 2016.

\bibitem{Charrow2015InformationTheoreticPW}
B.~Charrow, G.~Kahn, S.~Patil, S.~Liu, K.~Goldberg, P.~Abbeel, N.~Michael, and
  V.~R. Kumar, ``Information-theoretic planning with trajectory optimization
  for dense 3d mapping,'' \emph{Robotics: Science and Systems XI}, 2015.

\bibitem{mannucci2018}
A.~Mannucci, S.~Nardi, and L.~Pallottino, ``{Autonomous 3D Exploration of Large
  Areas: A Cooperative Frontier-Based Approach"},'' in \emph{Modelling and
  Simulation for Autonomous Systems}.\hskip 1em plus 0.5em minus 0.4em\relax
  Cham: Springer International Publishing, 2018.

\bibitem{Colares2016TheNF}
R.~G. Colares and L.~Chaimowicz, ``The next frontier: combining information
  gain and distance cost for decentralized multi-robot exploration,''
  \emph{Proceedings of the 31st Annual ACM Symposium on Applied Computing},
  2016.

\bibitem{Hardouin2020}
G.~Hardouin, J.~Moras, F.~Morbidi, J.~Marzat, and E.~M. Mouaddib,
  ``{Next-Best-View planning for surface reconstruction of large-scale 3D
  environments with multiple UAVs},'' in \emph{2020 IEEE/RSJ International
  Conference on Intelligent Robots and Systems (IROS)}, 2020.

\bibitem{Dutta2020}
A.~Dutta, A.~Bhattacharya, O.~P. Kreidl, A.~Ghosh, and P.~Dasgupta,
  ``{Multi-robot informative path planning in unknown environments through
  continuous region partitioning},'' \emph{International Journal of Advanced
  Robotic Systems}, 2020.

\bibitem{morilla2022sweep}
D.~Morilla-Cabello, L.~Bartolomei, L.~Teixeira, E.~Montijano, and M.~Chli,
  ``{Sweep-Your-Map: Efficient Coverage Planning for Aerial Teams in
  Large-Scale Environments},'' \emph{IEEE Robotics and Automation Letters},
  2022.

\bibitem{Zhou2019robust}
B.~Zhou, F.~Gao, L.~Wang, C.~Liu, and S.~Shen, ``{Robust and Efficient
  Quadrotor Trajectory Generation for Fast Autonomous Flight},'' \emph{IEEE
  Robotics and Automation Letters}, 2019.

\bibitem{oleynikova2016continuous}
H.~Oleynikova, M.~Burri, Z.~Taylor, J.~Nieto, R.~Siegwart, and E.~Galceran,
  ``{Continuous-Time Trajectory Optimization for Online UAV Replanning},'' in
  \emph{IEEE/RSJ International Conference on Intelligent Robots and Systems
  (IROS)}, 2016.

\bibitem{bartolomei2022multi}
L.~Bartolomei, Y.~Kompis, L.~Teixeira, and M.~Chli, ``{Autonomous Emergency
  Landing for Multicopters using Deep Reinforcement Learning},'' in \emph{2022
  {IEEE/RSJ} International Conference on Intelligent Robots and Systems
  ({IROS})}, 2022.

\end{thebibliography}

\vfill

\end{document}